\pdfoutput=1

\documentclass[11pt]{article}

\usepackage{acl}
\usepackage{pgfplots}
\usetikzlibrary{patterns}
\usepackage{amssymb}
\usepackage{bm}
\usepackage{times}

\usepackage{latexsym}
\usepackage{amsmath}
\usepackage{bm}
\usepackage{algpseudocode}
\usepackage{algorithm}
\usepackage{graphicx}
\usepgfplotslibrary{dateplot}
\usepackage{booktabs}
\usepackage{subcaption} 

\usepackage[T1]{fontenc}

\usepackage[utf8]{inputenc}

\usepackage{microtype}

\title{Blending Is All You Need: Cheaper, Better Alternative to Trillion-Parameters LLM}

\author{
Xiaoding Lu$^\S$
\quad
Zongyi Liu
\quad
Adian Liusie$^\S$
\quad
Vyas Raina$^\S$
\quad
Vineet Mudupalli$^\S$
\AND
Yuwen Zhang$^\P$
\quad
William Beauchamp$^\S$
\\\\
$^\S$University of Cambridge
\quad
$^\P$University College London
\\\\
Chai Research
}

\begin{document}
\maketitle
\begin{abstract}

In conversational AI research, there's a noticeable trend towards developing models with a larger number of parameters, exemplified by models like ChatGPT. While these expansive models tend to generate increasingly better chat responses, they demand significant computational resources and memory. This study explores a pertinent question: Can a combination of smaller models collaboratively achieve comparable or enhanced performance relative to a singular large model? We introduce an approach termed \textit{Blending}, a straightforward yet effective method of integrating multiple chat AIs. Our empirical evidence suggests that when specific smaller models are synergistically blended, they can potentially outperform or match the capabilities of much larger counterparts. For instance, integrating just three models of moderate size (6B/13B parameters) can rival or even surpass the performance metrics of a substantially larger model like ChatGPT (175B+ parameters). This hypothesis is rigorously tested using A/B testing methodologies with a large user base on the Chai research platform over a span of thirty days. The findings underscore the potential of the \textit{Blended} strategy as a viable approach for enhancing chat AI efficacy without a corresponding surge in computational demands.~\footnote{All trained models are provided at \url{https://huggingface.co/ChaiML}.}

\end{abstract}

\section{Introduction}
Due to the remarkable capabilities of current generative AI technology, pre-trained large language models (LLMs) have found extensive utilization across a diverse array of applications. One such application is in chat AI, where automatic systems are deployed as conversational assistants that keep users engaged in entertaining conversations. A common finding is that as one scales up the number of model parameters and the training data size, the quality and ability of the LLMs dramatically increases. This has led to the current trend of scaling up models to tremendous sizes, with current state-of-the-art systems having hundreds of billions of parameters. Although this has enabled highly capable chat AI with extraordinary emergent abilities, this comes at a practical cost of large inference overheads, with specialized infrastructure required, and access to these systems restricted through public APIs. It is therefore highly desirable to overcome these dramatic practical limitations, and have smaller and efficient chat AIs, while keeping users engaged and maintaining the conversational quality that current 100B+ parameters LLMs have achieved. 

Although a single small model is unlikely to compete against the current behemoth state-of-the-art  LLMs, one may question whether a group of moderately-sized LLMs can together form a chat AI of equivalent or perhaps better ability. In this work, we introduce Blended, an innovative and simple approach where we demonstrate that, surprisingly, if responses are selected randomly from a group of base chat AIs, the resulting combined chat AI is highly capable and engaging, and can outperform systems with orders of magnitude more parameters. We interestingly observe that the blended model appears to take characteristics that are the ``best of all", and that by conditioning a response on the conversational history, a single model with particular properties learns abilities from other systems. This leads to more captivating and diverse responses, and a more engaging user experience. We demonstrate the effectiveness of Blended over large-scale A/B tests on real users on the CHAI platform, where our results show that a Blended ensemble with three 6-13B parameter LLMs, outcompetes OpenAI's 175B+ parameter ChatGPT. We observe significantly higher user retention for blended ensembles than for ChatGPT-based chat AIs, illustrating that users find Blended chat AIs to be more engaging, entertaining and useful, despite Blended only requiring a fraction of the inference cost and memory overhead.

\section{Related Work}

\subsection{Chat AI approaches}
Chat AIs have been developed for a variety of applications, from user assistance to casual interactions (for \textit{chitchat})~\cite{10.1145/3166054.3166058}. Early designs were based on rule-based algorithms~\cite{10.1145/365153.365168} which later progressed to generative retrieval-based models~\cite{nlp4convai-2021-natural}. The emergence of pre-trained transformer language models marked a significant change in chat AI development~\cite{zhu2022simple, NIPS2017_3f5ee243, 10.1145/3373017.3373028}, where scaling-up trends led to increasingly larger Transformer-based models finetuned to conversational datasets for the development of chat AIs~\cite{ DBLP:journals/corr/abs-2001-09977, roller-etal-2021-recipes, DBLP:journals/corr/abs-2006-16779, choudhary-kawahara-2022-grounding, yan2022deep}.

Traditionally, chat AIs have been trained with self-supervised methods on conversational datasets. However, more recent approaches highlight the importance of human feedback in training to align better with human expectations of an engaging conversation~\citep{DBLP:journals/corr/abs-1811-07871, DBLP:journals/corr/abs-2112-00861, DBLP:journals/corr/abs-2001-09768}. This is typically achieved through either reinforcement learning from human feedback \citep[RLHF;][]{christiano2017deep, stiennon2020learning} or by using the reward model on its own to select or filter out responses \citep{DBLP:journals/corr/abs-1912-02164, irvine2023rewarding}

In our work, our \textit{Blended} approach does not consider how one can train better conversational LLMs, and instead demonstrates that one can leverage a group of existing small conversational LLMs and encourage them to collaborate over a conversation to form a single chat AI that generates more engaging and diverse responses.  

\subsection{Generative system combination}

Systems combination has been well-explored for deep-learning systems, with approaches such as stacking \cite{wolpert1992stacked}, negative correlation learning \cite{liu1999ensemble}, max-voter schemes \cite{ju2018relative, simonyan2014very} or probability averaging \cite{he2016deep, raina_gales_knill_2020, szegedy2015going} employed for a range of regression and classification tasks. With these ensembling methods, it has further been shown that increasing the diversity of the individual members can lead to better-performing combined systems \cite{kilimci2018deep, seijo2017ensemble}.

However, for generative language tasks where the outputs are a sequence of tokens, most ensembling approaches become inapplicable and ineffective. Sequence-level ensembling approaches, though, get around this by often averaging conditional token level probabilities of multiple systems \cite{sennrich2015improving, freitag2017ensemble, malinin2021uncertainty, gec-ensemble-dist}. This approach, however, often requires identical member architectures and access to the output probabilities of the tokens. With an increasing trend of limited black box access to LLMs (e.g. ChatGPT~\cite{liu2023summary} and BARD \cite{Nyberg_2021}), ensembling methods that only use output sequences may have practical benefit. Minimum Bayes' Risk (MBR) decoding~\citep{kumar-byrne-2004-minimum} enables this by using system outputs to select the predicted `best' system output. Though this approach has traditionally been used for Automatic Speech Recognition (ASR), it has also been successfully applied to NLP tasks ~\citep{rosti-etal-2007-combining, freitag-etal-2022-high, manakul2023cued, raina2023minimum}. With a growing number of (API-access only) deployed large language models, performing well at different tasks, \cite{jiang2023llmblender} also observed the need for a method to combine outputs in a blackbox setting. They propose \textit{LLM-Blender} to \textit{blend} the outputs from different language models by first ranking the outputs as per a \textit{PairRanker} and then \textit{fuse} the top-K outputs using a separate deep sequence-to-sequence system (termed \textit{GenFuser}).

As with MBR and LLM-Blender, in this work we also propose an ensembling approach that is able to combine outputs from blackbox language models. However, by designing our method for the specific nature of a multi-turn task (such as dialogue agents) our Blended approach does not require all component systems to generate outputs but instead stochastically selects the system that generates the next response, allowing for model blending at the level of a multi-turn conversation.

\section{Blended}
\subsection{Chat AI}
The objective of a chat AI is to design an automatic system that can produce engaging and entertaining conversations that human users can interact with. Let $u_k$ denote the user's $k$th turn, where each user turn is a sequence of words, $u_k\!=\!(w^{(k)}_1\hdots, w^{(k)}_{|u_k|})$. Similarly, let $r_k$ denote the system's $k$th generated response, which is also a sequence of words $r_k\!=\!(w^{(k)}_1, \hdots, w^{(k)}_{|r_k|})$. As an implicit language model, a particular chat AI, parameterised by $\theta$, models the probability of the next response given the previous conversational history,
\begin{equation}
    P(r_k | u_{1:k}, r_{1:k-1}; \theta)
\end{equation}
During training, the system implicitly learns to assign higher probability to responses that are fluent, engaging and high quality. Therefore an output can simply be sampled from its distribution, either stochastically, or through an approximate search process such as beam search. 
\begin{equation} \label{eqn:response}
    r_k \sim P(r | u_{1:k}, r_{1:k-1}; \theta)
\end{equation}
Inspired by InstructGPT~\cite{ouyang2022training} and outlined in \cite{irvine2023rewarding}, state-of-the-art chat AIs tends to follow a three-stage-pipeline. First, a pre-trained language model (PrLM) is fine-tuned on a relevant textual domain, e.g. entertaining literature for the design of an \textit{engaging} chatbot. Second, a reward model is trained using explicit human feedback, for example, by using user engagement as a proxy for response quality \citep{irvine2023rewarding}. Then finally, the reward model is used to improve the original PrLM, either by Proximal Policy Optimisation \cite{ouyang2022training} or by following a simple rejection sampling strategy. 

In developing a particular chat AI, there are many design choices such as the base PrLM, the conversational data used in fine-tuning, and the nature of human feedback used to update the system. One may expect that different recipes and training seeds may lead to highly diverse systems that each demonstrate unique strengths and characteristics. One can then consider how a set of chat AIs can be combined for a system with overall better characteristics.


\subsection{Ensembling}
In accordance with Bayesian statistical principles, the probability assigned to a particular response can be conceptualized as the marginal expectation taken over all plausible chat AI parameters,
\begin{align}
P(r_k | &u_{1:k}, r_{1:k-1}) \\
=& \mathbb E_{\theta\sim P_{\Theta}}\left[ P(r_k | u_{1:k}, r_{1:k-1}; \theta) \right ] \\
=& \int P_{\Theta}(\theta)P(r_k | u_{1:k}, r_{1:k-1}; \theta) d\theta
\end{align}
In practice, where we only have access to a finite set of chat AI systems $\{\theta_1, \theta_2... \theta_N \}$, one can approximate the continuous integral as a discrete summation. Further, one can assume that $P_{\Theta}(\theta)$ is distributed uniformly over the systems such that $P_{\Theta}(\theta_n) = \frac{1}{N}$, which may be a valid assumption if the set consists of similarly performing models. This yields the approximation,
\begin{align} \label{eqn:blend}
    \;P(r_k | &u_{1:k}, r_{1:k-1}) \\
    \approx& \sum_{\theta}P_{\Theta}(\theta)P(r_k | u_{1:k}, r_{1:k-1}; \theta) \\
    =& \frac{1}{N}\sum_{n=1}^NP(r_k | u_{1:k}, r_{1:k-1}; \theta_n)
    \label{eq:blended_theory}
\end{align}


\subsection{Blended}
The objective of our approach is to approximately draw samples from the true ensemble distribution (equation \ref{eq:blended_theory}). To achieve this approximation, each turn Blended randomly (and uniformly) selects the chat AI $\theta$ that generates the current response. This process is illustrated in Algorithm \ref{alg:blended}. It can be noted that during a conversation, the response generated by a specific chat AI is conditional on all previous responses generated by the previously selected chat AIs. This means that the different chat AIs are able to implicitly influence the output of the current response. As a result, the current response is a \textit{blending} of individual chat AI strengths, as they \textit{collaborate} to create an overall more engaging conversation. 

\begin{algorithm} 
\caption{Blended Algorithm}
\begin{algorithmic}[1]
\State $k \gets 1$
\While{true}
    \State $ u_k \gets \text{user's current input turn}$
    \State Sample model parameter $\theta_n\sim P_{\Theta}$
    \State Generate response \( r_k \) according to:
    \[ r_k \sim P(r | u_{1:k}, r_{1:k-1}; \theta_n) \]
    \State \( k = k + 1 \)
\EndWhile
\end{algorithmic}
\label{alg:blended}
\end{algorithm}


\label{sec:eval}
\section{Evaluating Chat AIs}
Evaluating the quality of NLG outputs is a notoriously challenging task \cite{fabbri2021summeval, liusie2023llm}, where traditional gold-standard approaches use human evaluators that score the quality of generated responses, which can be costly. However, since chat AIs are by definition deployed in social environments with humans, one can leverage statistics of users interaction as a meaningful and aligned measure of chat AI engagingness and quality. To assess the 'quality' of a chat AI, we consider two main proxy functions: the industry standard \textit{user retention} and the main objective function, \textit{user engagement}. 

\subsection{User Retention}
User retention is a standard industrial measure of a platform's success by measuring the fraction of users that return to the platform $k$ days after joining. Let the control group $\mathcal{G}_n$ be a randomly selected group of new users, where each user in this group will only be served chat AI $\theta_n$. Let $S_n(k)$ be the number of users from $\mathcal{G}_n$ that use the platform and interact with the chat AI on day $k$. Therefore, the $k$-day user retention rate, $R(k)$, is simply given by the fraction,
\begin{equation} \label{eqn:retention}
    R(k) = \frac{S_n(k)}{|\mathcal{G}_n|}.
\end{equation}
Retention rates from different models can be compared throughout the A/B testing period, where one can compare the immediate and long-term engagement of different chat AIs. Hence, for a considered group $\mathcal G_n$ and control group $\mathcal G_c$, one can define the test to control \textbf{retention ratio}, $q_n(k)$ as
\begin{equation}\label{eqn:ret-ratio}
    q_n(k) = \frac{R_n(k)}{R_c(k)}.
\end{equation}
Beyond comparing models, it is useful to extract retention curve statistics that can summarize a chat AI's performance with interpretable metrics. Empirical evidence suggests that the retention rate can be modelled well as,
\begin{equation}\label{eqn:retention-opt}
    R^*(k) = \frac{R(1)}{k^{-\beta}},
\end{equation}
where the parameter $\beta$ indicates the rate of user retention decay days, $k$. Taking the log of both sides yields;
\begin{align} \label{eqn:eng-ratio-opt}
    \log(q^*(k)) = \Delta\zeta + \Delta\beta \log k,
\end{align}
where $\Delta\zeta = (\log(R_w(1)) - \log(R_c(1))$ and $\Delta\beta=(\beta_w - \beta_c)$. One can therefore use the gradient and intercept of the log-log linear best-fit line to estimate the parameters $\Delta\beta$ and $\Delta\zeta$, which gives a useful comparison of the initial retention ratio and retention ratio decay rate relative to the control chat AI.

\subsection{User Engagement}
User retention is a useful industry metric, however, it may not perfectly align with the metrics that are of true interest. High-quality, engaging conversations are likely to keep users captivated for longer; therefore we directly define a proxy user engagement metric as the average time spent per visiting user. Let $E^{(u)}(t)$ represent whether a user is \textit{engaged} at a time $t$,
\begin{equation}
     E^{(u)}(t) = \begin{cases} 1, \hspace{0.6em} \text{user interacts in }t-\Delta \text{ to }t+\Delta,\\
     0, \hspace{0.6em}\text{otherwise},
     \end{cases}
\end{equation}
Then we can define $E_n(t)$, the engagement at time $t$ for all users in cohort $\mathcal{G}_n$, as
\begin{equation}
    E_n(t) = \frac{1}{|\mathcal G_n|} \sum_{u\in\mathcal{G}_n} E^{(u)}(t).
\end{equation}
As with user retention, the A/B setting allows for direct comparison of the engagement between different chat AIs. Hence we define the test to control \textbf{engagement ratio}, $r_n(t)$ as
\begin{equation} \label{eqn:eng-ratio}
    r_n(t) = \frac{E_n(t)}{E_c(t)}.
\end{equation}
It is also useful to have an overall single metric for the engagement score of a chat AI over time $t$. Hence, to obtain this, it is empirically observed that a sensible approximation for a chat AI engagement's decay is~\footnote{Periodic oscillations are not modeled here.},
\begin{equation}\label{eqn:eng-model}
    E^*(t) = \alpha t^{\gamma},
\end{equation}
This then gives a model for the test to control engagement ratio as
\begin{align} \label{eqn:eng-ratio-opt}
    \log(r^*(t)) &= \Delta\alpha + \Delta\gamma \log t,
\end{align}
where $\Delta\alpha=(\log(\alpha^{(w)})-\log(\alpha^{(c)}))$ and $\Delta\gamma=(\gamma^{(w)} - \gamma^{(c)}))$. By plotting $r(t)$ against $t$, a linear line of best fit can be found, with the parameters $\Delta\alpha$ and $\Delta\gamma$ being the intercept and gradient respectively. This gives the summarising metrics $\Delta\alpha$ and $\Delta\gamma$ to compare the engagement quality of different test chat AIs.

\section{Experiments}

\begin{figure}[h]
\centering
\begin{tikzpicture}
    \begin{axis}
        [
            ybar,
            bar width=5pt,
            legend style={at={(0.5, 0.97)},
            anchor=north,legend columns=-1},
            ylabel={Improvement Over Control \%},
            ylabel style={yshift=-10pt, inner sep=-1pt},
            symbolic x coords={1,2,3,4},
            xtick=data,
            xticklabels={{Blend (13,6,6B)}, GPT3.5 (175B), Vicuna+ (13B), ChaiLLM (6B)},
            xticklabel style={rotate=80, anchor=east},
            xtick pos=left,
            ymajorgrids=true,
            width=0.5\textwidth,
            ymin=0,
            ymax=120,
        ]
        \addplot[pattern=north west lines, pattern color=black, area legend] coordinates {(1, 110) (2, 82.5) (3, 22) (4, 23.2)};
        \addplot[fill=white, area legend] coordinates {(1, 71.2) (2, 20.5) (3, 10.2) (4, 10.1)};
        \legend{Engagement,Retention}
    \end{axis}
\end{tikzpicture}
\caption{Model performance comparisons, setting the baseline as Pygmalion 6B. Each model is assigned to 5,000 unique new users, graphs report the day 30 retention and engagement improvement with respect to the baseline.}
\label{fig:performance}
\end{figure}

\subsection{Experimental Set Up}
\noindent\textbf{Base chat AI systems:} In our experiments we consider four different base chat AI systems. We first have 3 moderately sized open-sourced LLMs: Pygmillion 6B\footnote{\url{https://huggingface.co/PygmalionAI/pygmalion-6b}}, Chai Model 6B\footnote{\url{https://huggingface.co/ChaiML/edit_sft_pyg_v2e_cp_17515}} and Vicuna 13B\footnote{\url{https://huggingface.co/lmsys/vicuna-13b-v1.3}}. Each base LLM has been further finetuned on conversational data, and uses rejection sampling from a trained reward model (detailed in \cite{irvine2023rewarding}). We finally also consider the state of art chat AI,  OpenAI's Davinci (GPT3.5), which has 175B parameters and is only available through a closed API call.  \\

\noindent\textbf{Methodology:} Each of the base chat AI systems are deployed with A/B tests on independent user groups, as discussed in Section \ref{sec:eval}, where the groups are of real users engaging with the Chai Research Platform. We conduct a large-scale evaluation with at least 10000 users in each group, and we monitor the user engagement on the platform over a 30-day period. Further, we deploy our blended system (Blended), encompassing Pygmillion, Chai Model and Vicuna. Since there can be external factors that may influence users' retention and engagement (e.g. platform popularity, holidays etc.), systems are only compared using relative engagement and relative retention, which are the metrics normalised to the selected baseline group. 
\begin{figure*}[t]
    \centering
    \begin{minipage}{0.48\textwidth}
        \centering
        \begin{tikzpicture}[scale=0.92]
        \begin{axis}[
            title={Engagement vs Inference Speed},
            xlabel={Relative Inference Speed (1/FLOPs)},
            ylabel={Improvement Over Baseline},
            ylabel style={yshift=-5pt, inner sep=-1pt},
            xmin=0, xmax=1.,
            ymin=0.1, ymax=1.4,
            grid=both,
            scatter/classes={
                a={mark=*,draw=black}
            }
        ]
        
        \addplot[
            scatter,
            only marks,
            scatter src=explicit symbolic,
            mark options={scale=1.3}
            ]table[meta=label] {
        x       y       label
        0.1     0.82    a
        0.72    1.15     a
        0.35    0.25    a
        1.0    0.25     a
        };
        
        \addplot[smooth, thick, gray, line width=2pt] coordinates {
            (0 ,  0.7)
            (0.2, 0.64)
            (0.4, 0.58)
            (0.6, 0.52)
            (1.0, 0.4)
        };
        
        \node at (axis cs:0.2,0.85) [above] {GPT3.5 (175B)};
        \node at (axis cs:0.72,1.2) [above] {Blend (13,6,6B)};
        \node at (axis cs:0.35,0.25) [above] {Vicuna+ (13B)};
        \node at (axis cs:0.825, 0.25) [above] {ChaiLLM (6B)};
        
        \end{axis}
    \end{tikzpicture}
    \caption{User Engagement}
    \label{fig:eng-scatter}
    \end{minipage}
    \hfill
    \begin{minipage}{0.48\textwidth}
        \centering
        \begin{tikzpicture}[scale=0.92]
        \begin{axis}[
            title={Retention vs Inference Speed},
            xlabel={Relative Inference Speed (1/FLOPs)},
            ylabel={Improvement Over Baseline},
            ylabel style={yshift=-5pt, inner sep=-1pt},
            xmin=0, xmax=1.,
            ymin=0., ymax=0.9,
            grid=both,
            scatter/classes={
                a={mark=*,draw=black}
            }
        ]
        
        \addplot[
            scatter,
            only marks,
            scatter src=explicit symbolic,
            mark options={scale=1.3}
            ]table[meta=label] {
        x       y       label
        0.1     0.205    a
        0.72    0.712    a
        0.35    0.102    a
        1.0     0.101    a
        };
        
        \addplot[smooth, thick, gray, line width=2pt] coordinates {
            (0 ,  0.17)
            (0.2, 0.146)
            (0.4, 0.122)
            (0.6, 0.098)
            (1.0, 0.05)
        };
        
        \node at (axis cs:0.2,0.235) [above] {GPT3.5 (175B)};
        \node at (axis cs:0.72,0.742) [above] {Blend (13,6,6B)};
        \node at (axis cs:0.38,0.132) [above] {Vicuna+ (13B)};
        \node at (axis cs:0.825, 0.131) [above] {ChaiLLM (6B)};
        
        \end{axis}
    \end{tikzpicture}
    \caption{User Retention}
    \label{fig:ret-scatter}
    \end{minipage}
\end{figure*}

\subsection{Experimental Results}
For each chat AI deployed on the Chai Research platform, we compute the user engagement for each day $k$, as per Equation \ref{eqn:eng-ratio} in an A/B test setting. By considering the 20th day ($k=20$), Figure \ref{fig:performance}a shows the engagement ratio of Blended, its constituent chat AIs and Open AI's GPT-3.5. We observe that the moderate-sized chat AIs (Pygmillion, Vicuna and ChaiLLM) have significantly lower engagement than that of GPT3.5, which is expected as GPT3.5 has over an order of magnitude more parameters. However, by blending the three base chat AIs, not only does Blended have higher engagement than each of the constituent systems, but the performance gains are so significant that Blended can outperform OpenAI's GPT3.5. The success of Blended over other chat AIs can also be observed when comparing the $k=20$ user retention ratio (Equation \ref{eqn:ret-ratio}), as seen in Figure \ref{fig:performance}.

We highlight that Blended has a total of 25B parameters compared to OpenAIs 175B parameters, and further, since responses for Blended are each sampled from a single component chat AI, the inference cost is equivalent to that of a single 6B/13B system. The significant difference in inference speed (measured as the inverse of total Floating Point Operations at test time) is highlighted in Figures \ref{fig:eng-scatter} and \ref{fig:eng-scatter} respectively, where it can be observed that Blended offers significant performance gains with respect to engagement and user retention, with speeds similar to that of small chat AIs. Implications of this are strong: instead of scaling up systems to improve quality, one can simply blend multiple smaller open-source systems, and without increasing any inference costs can drastically improve a user's conversational experience. This demonstrates the importance of model collaboration over simple model parameter scaling when designing engaging and successful chat AIs.

As an objective comparison, Table \ref{tab:stats} reports the single metric summaries (proposed in Section \ref{sec:eval}). With Pygmillion as the control, we report the test-to-control engagement ratio metrics $\Delta\alpha$ and $\Delta\gamma$, as well as the test-to-control retention ratio metrics $\Delta\zeta$ and $\Delta\beta$. Blended has the highest relative initial engagement, $\Delta\alpha$ and the best engagement ratio decay rate, $\Delta\gamma$. Although the \textit{retention} ratio decay rate, $\Delta\beta$ is better for Vicuna than Blended, Vicuna has a significantly lower initial retention ratio, $\Delta\zeta$, demonstrating that Vicuna would require an extended period of time to reach Blended's retention score~\footnote{This period of time is estimated to be around one year.}, as can be seen from figures \ref{fig:ret-scatter}. Overall it is clear that Blended, using a collaboration of smaller chat AIs, is effective in offering higher quality conversations than a single, much larger chat AI (OpenAI's GPT3.5).

\begin{table}[htb!]
    \centering
    \small
    \begin{tabular}{l|cc|cc|c}
    \toprule
    chat AI & $\Delta\zeta$  & $\Delta\beta$  & $\Delta\gamma$ &$\Delta\alpha$ & FLOP\\ \midrule
       Chai  & 0.1 & 0.0&0.3&0.2 & 1.0\\
        Vicuna  & -0.4 &0.9&0.0&0.1 & 2.2\\
        Pygmillion (\textbf{ctrl}) & 0.0&0.0&0.0&0.0&1.0\\ \midrule
        Blended & \underline{\textbf{0.2}}&\underline{0.5}&\underline{\textbf{2.1}}&\underline{\textbf{1.7}} & \underline{1.4}\\
        GPT3.5  & 0.0 & 0.3& 1.4& 0.5 & 29.2\\
        \bottomrule
    \end{tabular}
    \caption{Test to Control Retention and Engagement summary statistics and inference time (total Floating Point Operations / control) for component chat AIs (ChaiModel, Vicuna, Pygmillion (\textbf{control}); Blended and OpenAI's Davinci GPT3.5.}
    \label{tab:stats}
\end{table}

\section{Future Work}

The work demonstrated that Blended, a collaboration of multiple small chat AIs, performs better than a single large-scale chat AI, such as OpenAI's Davinci (ChatGPT). In this section we offer methods by which the Blended model can be further improved to create even more engaging user conversations.\newline  

\noindent\textbf{Selection set scaling}: Experiments in this work have demonstrated that with even a selection set of three component chat AIs (Chai model, Vicuna and Pygmillion), Blended is able to perform better than the much larger Davinci GPT3.5 model. This performance gain is attributed to the individual expertise of each individual component model that creates a conversation with a diverse set of qualities as the component systems collaborate. Hence, one simple approach to further increase the diversity and thus richness in the conversation is to scale to more than three component systems. Increasing the number of component systems has no computational cost, as inference is always only run through a single system for each response in Blended's methodology. Therefore, future work will explore the impact of increasing the selection set of component chat AIs on the overall quality of conversations.\newline

\noindent\textbf{Optimal Selection Distribution}: As demonstrated in Equation \ref{eqn:blend}, Blended in this work adopts a simple approximation for model selection, $P_{\Theta}(\theta_n) = \frac{1}{N}$. However, although each component chat AI, $\theta_n$, may have some value to add to an overall conversation, an equal contribution from each chat AI may not be the optimal setup. Hence, to combat this, a better approximation for the model selection distribution can be made with,
\begin{equation}
    P_{\Theta}(\theta_n) = \mathcal F(u_{1:k}, r_{1:k-1})_n,
\end{equation}
where $\mathcal F$ is a deep-learning classifier trained to predict the probability distribution over the chat AI selection set for identifying the $\theta_n$ to give the next most \textit{engaging} response $r_k$. This classifier can be trained using standard signals from Human-Feedback to identify effective and ineffective responses generated in conversations, e.g. if the user regenerated the response it is indicative of being an undesirable response. Future work will explore methodologies to design and train such a classifier, $\mathcal F$ to allow for a more optimal (aligned with user engagement) distribution, $P_{\Theta}$ to select the component chat AI for each response, $r_k$. A further advantage of this approach is that we can now add new chat AIs to the selection set, without the risk of damaging the performance of Blended, as the classifier learns to de-weigh the contribution from bad quality chat AIs.

\section{Conclusions}
This paper introduced Blended, a simple approach of combining multiple chat AIs by stochastically selecting responses from the different systems. Though simple, the approach is surprisingly powerful and enables a group of three 6-13B parameter models to achieve retention and engagement that is superior to that of the 175B ChatGPT. We demonstrate findings over large scale user A/B tests, which highlights that blending might be a promising solution to improve the quality of chat AIs, all while maintaining inference costs of smaller systems.


\bibliography{anthology,custom}
\bibliographystyle{acl_natbib}
\end{document}